\documentclass[letterpaper, 10 pt, conference]{ieeeconf}
\IEEEoverridecommandlockouts
\usepackage[vlined, ruled, boxed, linesnumbered]{algorithm2e}

\SetCommentSty{mycommfont}

\SetKwProg{Function}{}{}{}

\usepackage{graphicx}
\usepackage{graphics}
\usepackage{times}
\usepackage{amsmath}
\usepackage{amssymb}
\usepackage{float}
\usepackage{url}
\usepackage{subfigure}
\usepackage{caption}
\usepackage{multirow}
\usepackage{gensymb}
\usepackage{epstopdf}
\usepackage{wrapfig}
\usepackage{setspace} 
\usepackage[noend]{algpseudocode}
\usepackage{color}
\usepackage{cite}
\usepackage{booktabs}
\usepackage{stackengine}
\usepackage[font={small}]{caption}

\usepackage{tikz}
\usepackage{textcomp}
\usepackage{lipsum}

\newcommand{\bomega}{\boldsymbol{\omega}}
\newcommand{\bx}{\bold{x}}
\newcommand{\bb}{\bold{b}}
\newcommand{\bn}{\bold{n}}
\newcommand{\ba}{\bold{a}}
\newcommand{\bg}{\bold{g}}
\newcommand{\bv}{\bold{v}}
\newcommand{\bp}{\bold{p}}
\newcommand{\bR}{\bold{R}}

\newcommand{\bBW}{\bold{BW}}
\newcommand{\Dt}{\Delta t}

\newcommand{\bT}{\bold{T}}

\newcommand{\bd}{\bold{d}}
\newcommand{\bM}{\bold{M}}
\newcommand{\bpoint}{\bold{p}}

\newcommand{\textF}{\text{F}}
\newcommand{\primeF}{'\textF}

\newcommand{\bigbar}{\Big|}
\newcommand{\bfracNoLine}[2]{\genfrac{|}{|}{0pt}{0}{#1}{#2}}

\title{\LARGE \bf
LIO-SAM: Tightly-coupled Lidar Inertial Odometry via\\ Smoothing and Mapping
}

\author{Tixiao Shan, Brendan Englot, Drew Meyers, Wei Wang, Carlo Ratti, and Daniela Rus
\thanks{
\scriptsize{
T. Shan, D. Meyers, W. Wang, and C. Ratti are with the Department of Urban Studies and Planning, Massachusetts Institute of Technology, USA, {\tt\scriptsize \{shant, drewm, wweiwang, ratti\}@mit.edu}. \newline
\indent B. Englot is with the Department of Mechanical Engineering, Stevens Institute of Technology, USA, {\tt\scriptsize benglot@stevens.edu}. \newline
\indent T. Shan, W. Wang, and D. Rus are with the Computer Science \& Artificial Intelligence Laboratory, Massachusetts Institute of Technology, USA, {\tt\scriptsize \{shant, wweiwang, rus\}@mit.edu}.}%
}
}

\begin{document}

\maketitle
\thispagestyle{empty}
\pagestyle{empty}


\begin{abstract}
We propose a framework for tightly-coupled lidar inertial odometry via smoothing and mapping, LIO-SAM, that achieves highly accurate, real-time mobile robot trajectory estimation and map-building. LIO-SAM formulates lidar-inertial odometry atop a factor graph, allowing a multitude of relative and absolute measurements, including loop closures, to be incorporated from different sources as factors into the system.  
The estimated motion from inertial measurement unit (IMU) pre-integration de-skews point clouds and produces an initial guess for lidar odometry optimization. The obtained lidar odometry solution is used to estimate the bias of the IMU.
To ensure high performance in real-time, 
we marginalize old lidar scans for pose optimization, rather than matching lidar scans to a global map. Scan-matching at a local scale instead of a global scale significantly improves the real-time performance of the system, as does the selective introduction of keyframes, and an efficient sliding window approach that registers a new keyframe to a fixed-size set of prior ``sub-keyframes.''
The proposed method is extensively evaluated on datasets gathered from three platforms over various scales and environments.
\end{abstract}


\section{Introduction}

State estimation, localization and mapping are fundamental prerequisites for a successful intelligent mobile robot, required for feedback control, obstacle avoidance, and planning, among many other capabilities. Using vision-based and lidar-based sensing, great efforts have been devoted to achieving high-performance real-time simultaneous localization and mapping (SLAM) that can support a mobile robot's six degree-of-freedom state estimation. Vision-based methods typically use a monocular or stereo camera and triangulate features across successive images to determine the camera motion. Although vision-based methods are especially suitable for place recognition, 
their sensitivity to initialization, illumination, and range 
make them unreliable when they alone are used to support an autonomous navigation system. On the other hand, lidar-based methods are largely invariant to illumination change. Especially with the recent availability of long-range, high-resolution 3D lidar, such as the Velodyne VLS-128 and Ouster OS1-128, lidar becomes more suitable to directly capture the fine details of an environment in 3D space. Therefore, this paper focuses on lidar-based state estimation and mapping methods.


Many lidar-based state estimation and mapping methods have been proposed in the last two decades. Among them, the lidar odometry and mapping (LOAM) method proposed in \cite{LOAM-2017} for low-drift and real-time state estimation and mapping is among the most widely used. LOAM, which uses a lidar and an inertial measurement unit (IMU), achieves state-of-the-art performance and has been ranked as the top lidar-based method since its release on the KITTI odometry benchmark site \cite{KITTI-2012}. Despite its success, LOAM presents some limitations - by saving its data in a global voxel map, it is often difficult to perform loop closure detection and incorporate other absolute measurements, e.g., GPS, for pose correction. Its online optimization process becomes less efficient when this voxel map becomes dense in a feature-rich environment. LOAM also suffers from drift in large-scale tests, as it is a scan-matching based method at its core.


\begin{figure*}[ht]
	\centering
	\includegraphics[width=.9\textwidth]{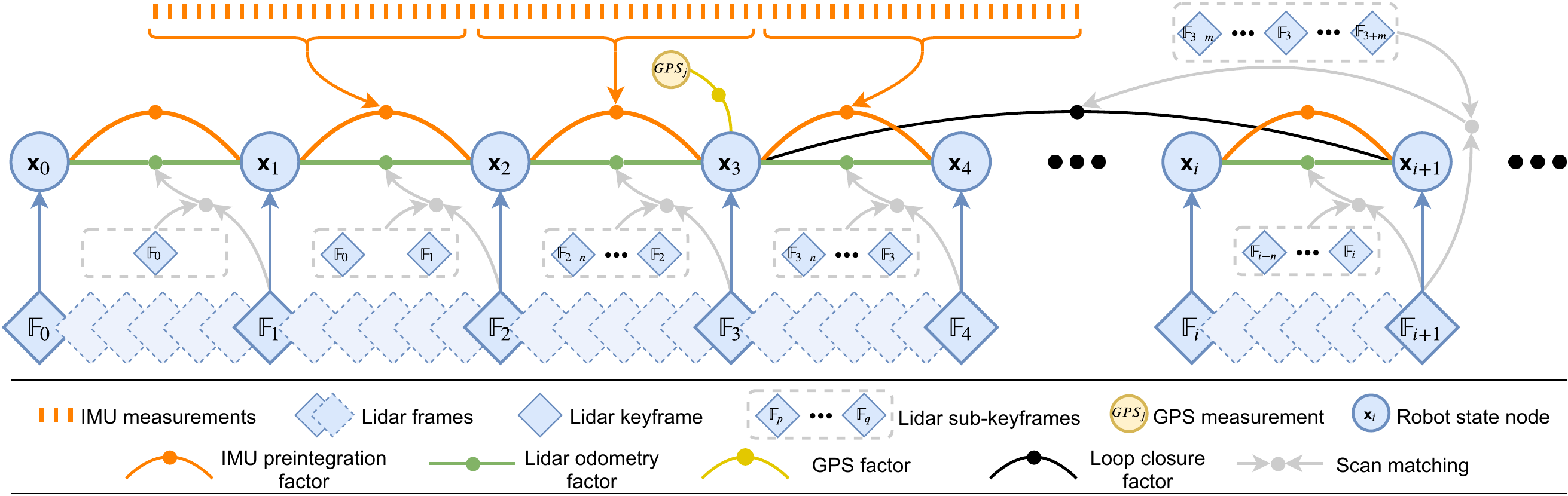}
	\caption{The system structure of LIO-SAM. The system receives input from a 3D lidar, an IMU and optionally a GPS. Four types of factors are introduced to construct the factor graph.: (a) IMU preintegration factor, (b) lidar odometry factor, (c) GPS factor, and (d) loop closure factor. The generation of these factors is discussed in Section \ref{sec::lio-sam}.}
	\label{fig::factor-graph}
	\vspace{-5mm}
\end{figure*}

In this paper, we propose a framework for tightly-coupled lidar inertial odometry via smoothing and mapping, LIO-SAM, to address the aforementioned problems. 
We assume a nonlinear motion model for point cloud de-skew, estimating the sensor motion during a lidar scan using raw IMU measurements. In addition to de-skewing point clouds, the estimated motion serves as an initial guess for lidar odometry optimization. The obtained lidar odometry solution is then used to estimate the bias of the IMU in the factor graph. 
By introducing a global factor graph for robot trajectory estimation, we can efficiently perform sensor fusion using lidar and IMU measurements,  incorporate place recognition among robot poses, and introduce absolute measurements, such as GPS positioning and compass heading, when they are available. This collection of factors from various sources is used for joint optimization of the graph. 
Additionally, we marginalize old lidar scans for pose optimization, rather than matching scans to a global map like LOAM. Scan-matching at a local scale instead of a global scale significantly improves the real-time performance of the system, as does the selective introduction of keyframes, and an efficient sliding window approach that registers a new keyframe to a fixed-size set of prior ``sub-keyframes.'' The main contributions of our work can be summarized as follows:
\begin{itemize}
	\item A tightly-coupled lidar inertial odometry framework built atop a factor graph, that is suitable for multi-sensor fusion and global optimization.
	\item An efficient, local sliding window-based scan-matching approach that enables real-time performance by registering selectively chosen new keyframes to a fixed-size set of prior sub-keyframes.
	\item The proposed framework is extensively validated with tests across various scales, vehicles, and environments.
\end{itemize}


\section{Related Work}

Lidar odometry is typically performed by finding the relative transformation between two consecutive frames using scan-matching methods such as ICP \cite{ICP-1992} and GICP \cite{GICP-2009}. Instead of matching a full point cloud, feature-based matching methods have become a popular alternative due to their computational efficiency. For example, in \cite{Finding-Plane-2013}, a plane-based registration approach is proposed for real-time lidar odometry. Assuming operations in a structured environment, it extracts planes from the point clouds and matches them by solving a least-squares problem. A collar line-based method is proposed in \cite{Collar-Line-2016} for odometry estimation. In this method, line segments are randomly generated from the original point cloud and used later for registration. However, a scan's point cloud is often skewed because of the rotation mechanism of modern 3D lidar, and sensor motion. Solely using lidar for pose estimation is not ideal since registration using skewed point clouds or features will eventually cause large drift. 

Therefore, lidar is typically used in conjunction with other sensors, such as IMU and GPS, for state estimation and mapping. Such a design scheme, utilizing sensor fusion, can typically be grouped into two categories: loosely-coupled fusion and tightly-coupled fusion. In LOAM \cite{LOAM-2017}, IMU is introduced to de-skew the lidar scan and give a motion prior for scan-matching. However, the IMU is not involved in the optimization process of the algorithm. Thus LOAM can be classified as a loosely-coupled method. A lightweight and ground-optimized lidar odometry and mapping (LeGO-LOAM) method is proposed in \cite{LEGOLOAM-2018} for ground vehicle mapping tasks \cite{Traversability-2018}. Its fusion of IMU measurements is the same as LOAM. A more popular approach for loosely-coupled fusion is using extended Kalman filters (EKF). For example, \cite{UAV-IROS-2013}-\cite{Hening-AIAA-2017} integrate measurements from lidar, IMU, and optionally GPS using an EKF in the optimization stage for robot state estimation. 

Tightly-coupled systems usually offer improved accuracy and are presently a major focus of ongoing research \cite{SLAM-Review-2018}. 
In \cite{IN2LAMA-ICRA-2019}, preintegrated IMU measurements are exploited for de-skewing point clouds. 
A robocentric lidar-inertial state estimator, R-LINS, is presented in \cite{Robocentric-2019}. R-LINS uses an error-state Kalman filter to correct a robot's state estimate recursively in a tightly-coupled manner. Due to the lack of other available sensors for state estimation, it suffers from drift during long-during navigation. 
A tightly-coupled lidar inertial odometry and mapping framework, LIOM, is introduced in \cite{LIO-MAPPING-2019}. LIOM, which is the abbreviation for LIO-mapping, jointly optimizes measurements from lidar and IMU and achieves similar or better accuracy when compared with LOAM. Since LIOM is designed to process all the sensor measurements, real-time performance is not achieved - it runs at about $0.6\times$ real-time in our tests. 

%


\section{Lidar Inertial Odometry via \\Smoothing and Mapping}
\label{sec::lio-sam}

\subsection{System Overview}
\label{sec::system-overview}

We first define frames and notation that we use throughout the paper. We denote the world frame as $\bold{W}$ and the robot body frame as $\bold{B}$. 
We also assume the IMU frame coincides with the robot body frame for convenience. The robot state $\bx$ can be written as:
\begin{align}
	\bx = {[\; \bR^{\bT}, \;\bp^{\bT}, \;\bv^{\bT}, \;\bb^{\bT} \;]}^{\bT} , 
\end{align}
where $\bR \in SO(3)$ is the rotation matrix, $\bp \in \mathbb{R}^{3}$ is the position vector, $\bv$ is the speed, and $\bb$ is the IMU bias. The transformation $\bT \in SE(3)$ from $\bold{B}$ to $\bold{W}$ is represented as $\bT = [\bR \;|\; \bp]$.

An overview of the proposed system is shown in Figure \ref{fig::factor-graph}. 
The system receives sensor data from a 3D lidar, an IMU and optionally a GPS. 
We seek to estimate the state of the robot and its trajectory using the observations of these sensors. This state estimation problem can be formulated as a maximum a posteriori (MAP) problem. We use a factor graph to model this problem, as it is better suited to perform inference when compared with Bayes nets. With the assumption of a Gaussian noise model, the MAP inference for our problem is equivalent to solving a nonlinear least-squares problem \cite{Dellaert-Factor-Graph-2017}. Note that without loss of generality, the proposed system can also incorporate measurements from other sensors, such as elevation from an altimeter or heading from a compass.

We introduce four types of $factors$ along with one $variable$ type for factor graph construction. This variable, representing the robot's state at a specific time, is attributed to the $nodes$ of the graph. The four types of factors are: (a) IMU preintegration factors, (b) lidar odometry factors, (c) GPS factors, and (d) loop closure factors. A new robot state node $\bx$ is added to the graph when the change in robot pose exceeds a user-defined threshold. The factor graph is optimized upon the insertion of a new node using incremental smoothing and mapping with the Bayes tree (iSAM2) \cite{iSAM2-2012}. The process for generating these factors is described in the following sections.
 

\subsection{IMU Preintegration Factor}
\label{sec::imu-factor}

The measurements of angular velocity and acceleration from an IMU are defined using Eqs. \ref{eq::imu-angular-velocity} and \ref{eq::imu-acceleration}:
\begin{align}
	\hat{\bomega}_{t} &= \bomega_{t} + \bb^{\bomega}_{t} + \bn^{\bomega}_{t} \label{eq::imu-angular-velocity} \\ 
	\hat{\ba}_{t} &= \bR^{\bBW}_{t} (\ba_{t} - \bg) + \bb^{\ba}_{t} + \bn^{\ba}_{t}, \label{eq::imu-acceleration}
\end{align}
where $\hat{\bomega}_{t}$ and $\hat{\ba}_{t}$ are the raw IMU measurements in $\bold{B}$ at time $t$. $\hat{\bomega}_{t}$ and $\hat{\ba}_{t}$ are affected by a slowly varying bias $\bb_{t}$ and white noise $\bn_{t}$. $\bR^{\bBW}_{t}$ is the rotation matrix from $\bold{W}$ to $\bold{B}$. $\bg$ is the constant gravity vector in $\bold{W}$.

We can now use the measurements from the IMU to infer the motion of the robot. The velocity, position and rotation of the robot at time $t+\Dt$ can be computed as follows:
\begin{align}
	\bv_{t+\Dt} &= \bv_{t} + \bg\Dt + \bR_{t}(\hat{\ba}_{t} - \bb^{\ba}_{t} - \bn^{\ba}_{t})\Dt \label{eq::velocity-prop}\\
	\begin{split}
	\bp_{t+\Dt} &=\bp_{t} + \bv_{t}\Dt + \frac{1}{2}\bg\Dt^{2} \label{eq::position-prop}\\ 
				  &\;\;\;\;\;\;\;\;\; + \frac{1}{2}\bR_{t}(\hat{\ba}_{t} - \bb^{\ba}_{t} - \bn^{\ba}_{t})\Dt^{2}
	\end{split}\\
	\bold{R}_{t+\Delta t} &= \bold{R}_{t}\;\bold{exp}((\hat{\bomega}_{t} - \bb^{\bomega}_{t} - \bn^{\bomega}_{t}) \Delta t), \label{eq::rotation-prop}
\end{align}
where $\bR_{t} = \bR^{\bold{WB}}_{t} = {\bR^{\bold{BW}}_{t}}^{\mathsf{T}}$. Here we assume that the angular velocity and the acceleration of $\bold{B}$ remain constant during the above integration.

We then apply the IMU preintegration method proposed in \cite{IMUPre-2016} to obtain the relative body motion between two timesteps. The preintegrated measurements $\Delta \bv_{ij}$, $\Delta \bp_{ij}$, and $\Delta \bR_{ij}$ between time $i$ and $j$ can be computed using:
\begin{align}
	\Delta \bv_{ij} &= \bR_{i}^{\mathsf{T}}(\bv_{j} - \bv_{i} - \bg\Dt_{ij}) \label{eq::delta-v}\\
	\Delta \bp_{ij} &= \bR_{i}^{\mathsf{T}}(\bp_{j} - \bp_{i} - \bv_{i}\Dt_{ij} - \frac{1}{2}\bg\Dt^{2}_{ij}) \label{eq::delta-p}\\
	\Delta \bR_{ij} &= \bR_{i}^{\mathsf{T}} \; \bR_{j}. \label{eq::delta-R}
\end{align}
Due to space limitations, we refer the reader to the description from \cite{IMUPre-2016} for the detailed derivation of Eqs. \ref{eq::delta-v} - \ref{eq::delta-R}.
Besides its efficiency, applying IMU preintegration also naturally gives us one type of constraint for the factor graph - IMU preintegration factors. The IMU bias is jointly optimized alongside the lidar odometry factors in the graph.


\subsection{Lidar Odometry Factor}
\label{sec::lidar-factor}

When a new lidar scan arrives, we first perform feature extraction. Edge and planar features are extracted by evaluating the roughness of points over a local region. Points with a large roughness value are classified as edge features. Similarly, a planar feature is categorized by a small roughness value. We denote the extracted edge and planar features from a lidar scan at time $i$ as $\textF^{e}_{i}$ and $\textF^{p}_{i}$ respectively. All the features extracted at time $i$ compose a lidar \textit{frame} $\mathbb{F}_{i}$, where $\mathbb{F}_{i} = \{\textF^{e}_{i}, \textF^{p}_{i}\}$. Note that a lidar frame $\mathbb{F}$ is represented in $\bold{B}$. A more detailed description of the feature extraction process can be found in \cite{LOAM-2017}, or \cite{LEGOLOAM-2018} if a range image is used.

Using every lidar frame for computing and adding factors to the graph is computationally intractable, so we adopt the concept of \textit{keyframe} selection, which is widely used in the visual SLAM field. 
Using a simple but effective heuristic approach, we select a lidar frame $\mathbb{F}_{i+1}$ as a keyframe when the change in robot pose exceeds a user-defined threshold when compared with the previous state $\bx_{i}$. The newly saved keyframe, $\mathbb{F}_{i+1}$, is associated with a new robot state node, $\bx_{i+1}$, in the factor graph. The lidar frames between two keyframes are discarded. Adding keyframes in this way not only achieves a balance between map density and memory consumption but also helps maintain a relatively sparse factor graph, which is suitable for real-time nonlinear optimization. In our work, the position and rotation change thresholds for adding a new keyframe are chosen to be $1m$ and $10^\circ$.

Let us assume we wish to add a new state node $\bx_{i+1}$ to the factor graph. The lidar keyframe that is associated with this state is $\mathbb{F}_{i+1}$. The generation of a lidar odometry factor is described in the following steps:

\subsubsection{Sub-keyframes for voxel map}
\label{sec::lidar-factor-map}
We implement a sliding window approach to create a point cloud map containing a fixed number of recent lidar scans. Instead of optimizing the transformation between two consecutive lidar scans, we extract the $n$ most recent keyframes, which we call the \textit{sub-keyframes}, for estimation. The set of sub-keyframes $\{\mathbb{F}_{i-n},...,\mathbb{F}_{i}\}$ is then transformed into frame $\bold{W}$ using the transformations $\{\bT_{i-n},...,\bT_i\}$ associated with them. 
The transformed sub-keyframes are merged together into a voxel map $\bM_{i}$. Since we extract two types of features in the previous feature extraction step, $\bM_{i}$ is composed of two sub-voxel maps that are denoted $\bM^{e}_{i}$, the edge feature voxel map, and $\bM^{p}_{i}$, the planar feature voxel map. The lidar frames and voxel maps are related to each other as follows:
\begin{align}
	\nonumber
	&\bM_{i} = \{\bM^{e}_{i}, \bM^{p}_{i}\} \\
	\nonumber
	&where: \bM^{e}_{i} =\; \primeF^{e}_{i} \cup \primeF^{e}_{i-1} \cup...\cup\primeF^{e}_{i-n} \\
	\nonumber
	&\;\;\;\;\;\;\;\;\;\;\;\;\; \bM^{p}_{i} =\; \primeF^{p}_{i} \cup \primeF^{p}_{i-1} \cup ... \cup \primeF^{p}_{i-n}.
\end{align}
$\primeF^{e}_{i}$ and $\primeF^{p}_{i}$ are the transformed edge and planar features in $\bold{W}$. $\bM^{e}_{i}$ and $\bM^{p}_{i}$ are then downsampled to eliminate the duplicated features that fall in the same voxel cell. In this paper, $n$ is chosen to be 25. The downsample resolutions for $\bM^{e}_{i}$ and $\bM^{p}_{i}$ are $0.2m$ and $0.4m$, respectively.

\subsubsection{Scan-matching}
\label{sec::scan-matching}

We match a newly obtained lidar frame $\mathbb{F}_{i+1}$, which is also $\{\textF^{e}_{i+1}, \textF^{p}_{i+1}\}$, to $\bM_{i}$ via scan-matching. Various scan-matching methods, such as \cite{ICP-1992} and \cite{GICP-2009}, can be utilized for this purpose. Here we opt to use the method proposed in \cite{LOAM-2017} due to its computational efficiency and robustness in various challenging environments. 

We first transform $\{\textF^{e}_{i+1}, \textF^{p}_{i+1}\}$ from $\bold{B}$ to $\bold{W}$ and obtain $\{ \primeF^{e}_{i+1}, \primeF^{p}_{i+1} \}$. This initial transformation is obtained by using the predicted robot motion, $\tilde{\bT}_{i+1}$, from the IMU. For each feature in $\primeF^{e}_{i+1}$ or $\primeF^{p}_{i+1}$, we then find its edge or planar correspondence in $\bM^{e}_{i}$ or $\bM^{p}_{i}$. For the sake of brevity, the detailed procedures for finding these correspondences are omitted here, but are described thoroughly in \cite{LOAM-2017}.

\subsubsection{Relative transformation}

The distance between a feature and its edge or planar patch correspondence can be computed using the following equations:
\begin{gather}
	\bd_{e_{k}} = \frac{\bigbar (\bpoint^{e}_{i+1, k} - \bpoint^{e}_{i, u}) \times (\bpoint^{e}_{i+1, k} - \bpoint^{e}_{i, v}) \bigbar}{\bigbar \bpoint^{e}_{i, u} - \bpoint^{e}_{i, v} \bigbar} \\
	\bd_{p_{k}} = \cfrac{ \bfracNoLine{(\bpoint^{p}_{i+1, k} - \bpoint^{p}_{i, u})}{(\bpoint^{p}_{i, u} - \bpoint^{p}_{i, v}) \times (\bpoint^{p}_{i, u} - \bpoint^{p}_{i, w})} }{\bigbar (\bpoint^{p}_{i, u} - \bpoint^{p}_{i, v}) \times (\bpoint^{p}_{i, u} - \bpoint^{p}_{i, w}) \bigbar},
\end{gather}
where $k$, $u$, $v$, and $w$ are the feature indices in their corresponding sets. For an edge feature $\bpoint^{e}_{i+1, k}$ in $\primeF^{e}_{i+1}$, $\bpoint^{e}_{i, u}$ and $\bpoint^{e}_{i, v}$ are the points that form the corresponding edge line in $\bM^{e}_{i}$. For a planar feature $\bpoint^{p}_{i+1, k}$ in $\primeF^{p}_{i+1}$, $\bpoint^{p}_{i, u}$, $\bpoint^{p}_{i, v}$, and $\bpoint^{p}_{i, w}$ form the corresponding planar patch in $\bM^{p}_{i}$.  
The Gauss–Newton method is then used to solve for the optimal transformation by minimizing:
\begin{align}
	\nonumber
    \min_{{\bT}_{i+1}}\bigg\{
        \sum_{\bpoint^{e}_{i+1,k} \in \primeF^{e}_{i+1}}{\bd_{e_{k}}} +
        \sum_{\bpoint^{p}_{i+1,k} \in \primeF^{p}_{i+1}}{\bd_{p_{k}}}
    \bigg\}.
\end{align}
At last, we can obtain the relative transformation $\Delta\bT_{i,i+1}$ between $\bx_{i}$ and $\bx_{i+1}$, which is the lidar odometry factor linking these two poses:
\begin{align}
	\Delta\bT_{i,i+1}=\bT_{i}^{\mathsf{T}}\bT_{i+1}
\end{align}

We note that an alternative approach to obtain $\Delta\bT_{i,i+1}$ is to transform sub-keyframes into the frame of $\bx_{i}$. In other words, we match $\mathbb{F}_{i+1}$ to the voxel map that is represented in the frame of $\bx_{i}$.	In this way, the \textit{real} relative transformation $\Delta\bT_{i,i+1}$ can be directly obtained. Because the transformed features $\primeF^{e}_{i}$ and $\primeF^{p}_{i}$ can be reused multiple times, we instead opt to use the approach described in Sec. \ref{sec::lidar-factor-map} for its computational efficiency. 

\subsection{GPS Factor}
\label{sec::gps-factor}

Though we can obtain reliable state estimation and mapping by utilizing only IMU preintegration and lidar odometry factors, the system still suffers from drift during long-duration navigation tasks. To solve this problem, we can introduce sensors that offer absolute measurements for eliminating drift. Such sensors include an altimeter, compass, and GPS. For the purposes of illustration here, we discuss GPS, as it is widely used in real-world navigation systems.

When we receive GPS measurements, we first transform them to the local Cartesian coordinate frame using the method proposed in \cite{ROBOT-LOCALIZATION}. Upon the addition of a new node to the factor graph, we then associate a new GPS factor with this node. If the GPS signal is not hardware-synchronized with the lidar frame, we interpolate among GPS measurements linearly based on the timestamp of the lidar frame.

We note that adding GPS factors constantly when GPS reception is available is not necessary because the drift of lidar inertial odometry grows very slowly. In practice, we only add a GPS factor when the estimated position covariance is larger than the received GPS position covariance.





\subsection{Loop Closure Factor}
\label{sec::loop-factor}

Thanks to the utilization of a factor graph, loop closures can also be seamlessly incorporated into the proposed system, as opposed to LOAM and LIOM. For the purposes of illustration, we describe and implement a naive but effective Euclidean distance-based loop closure detection approach. We also note that our proposed framework is compatible with other methods for loop closure detection, for example, \cite{LOOP-CLOSURE-2018} and \cite{LOOP-CLOSURE-2019}, which generate a point cloud descriptor and use it for place recognition.

When a new state $\bx_{i+1}$ is added to the factor graph, we first search the graph and find the prior states that are close to $\bx_{i+1}$ in Euclidean space. As is shown in Fig. \ref{fig::factor-graph}, for example, $\bx_{3}$ is one of the returned candidates. We then try to match $\mathbb{F}_{i+1}$ to the sub-keyframes $\{\mathbb{F}_{3-m}, ..., \mathbb{F}_{3}, ..., \mathbb{F}_{3+m}\}$ using scan-matching. Note that $\mathbb{F}_{i+1}$ and the past sub-keyframes are first transformed into $\bold{W}$ before scan-matching. We obtain the relative transformation $\Delta\bT_{3,i+1}$ and add it as a loop closure factor to the graph. Throughout this paper, we choose the index $m$ to be 12, and the search distance for loop closures is set to be $15m$ from a new state $\bx_{i+1}$.

In practice, we find adding loop closure factors is especially useful for correcting the drift in a robot's altitude, when GPS is the only absolute sensor available. This is because the elevation measurement from GPS is very inaccurate - giving rise to altitude errors approaching 100$m$ in our tests, in the absence of loop closures.  

  
\section{Experiments}

\begin{figure}[t!]
	\centering
	\subfigure[Handheld device]{\includegraphics[width=.32\columnwidth]{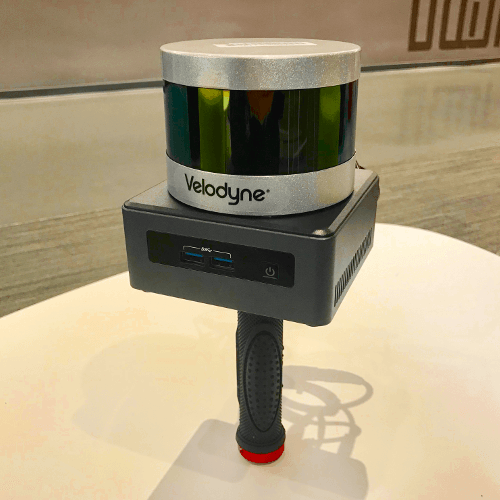}}
	\subfigure[Clearpath Jackal]{\includegraphics[width=.32\columnwidth]{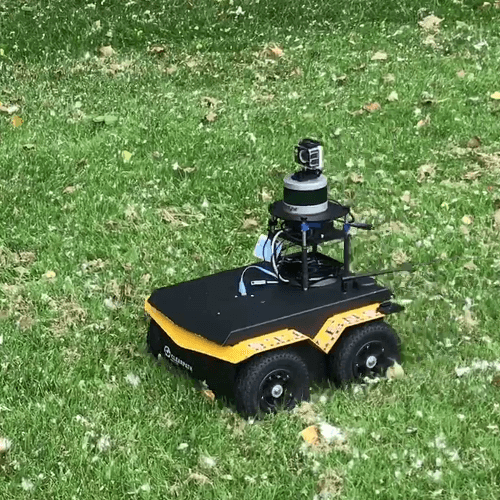}}
	\subfigure[Duffy 21]{\includegraphics[width=.32\columnwidth]{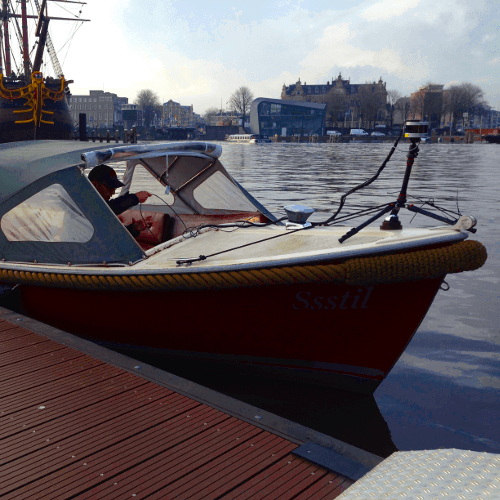}}
	\caption{Datasets are collected on 3 platforms: (a) a custom-built handheld device, (b) an unmanned ground vehicle - Clearpath Jackal, (c) an electric boat - Duffy 21.}
	\label{fig::experiments-equipments}
	\vspace{-3mm}
\end{figure}

We now describe a series of experiments to qualitatively and quantitatively analyze our proposed framework. The sensor suite used in this paper includes a Velodyne VLP-16 lidar, a MicroStrain 3DM-GX5-25 IMU, and a Reach M GPS. For validation, we collected 5 different datasets across various scales, platforms and environments. These datasets are referred to as \textit{Rotation}, \textit{Walking}, \textit{Campus}, \textit{Park} and \textit{Amsterdam}, respectively. The sensor mounting platforms are shown in Fig. \ref{fig::experiments-equipments}. The first three datasets were collected using a custom-built handheld device on the MIT campus. The Park dataset was collected in a park covered by vegetation, using an unmanned ground vehicle (UGV) - the Clearpath Jackal. The last dataset, Amsterdam, was collected by mounting the sensors on a boat and cruising in the canals of Amsterdam. The details of these datasets are shown in Table \ref{tab::dataset-details}.

\begin{table}[ht]
	\centering
	\caption{Dataset details}
	\label{tab::dataset-details}
	\resizebox{0.95\columnwidth}{!}{
		\begin{tabular}{@{}ccccc@{}}
			\toprule
			Dataset & Scans & \begin{tabular}[c]{@{}c@{}}Elevation\\ change (m)\end{tabular} & \begin{tabular}[c]{@{}c@{}}Trajectory\\ length (m)\end{tabular} & \begin{tabular}[c]{@{}c@{}}Max rotation\\ speed ($^{\circ}$/s)\end{tabular} \\ 
			\midrule
            Rotation & 582 & 0 & 0 & 213.9 \\
            Walking & 6502 & 0.3 & 801 & 133.7 \\
            Campus & 9865 & 1.0 & 1437 & 124.8 \\
            Park & 24691 & 19.0 & 2898 & 217.4 \\
            Amsterdam & 107656 & 0 & 19065 & 17.2 \\
			\bottomrule
		\end{tabular}
	}
\vspace{-3mm}
\end{table}

We compare the proposed LIO-SAM framework with LOAM and LIOM. In all the experiments, LOAM and LIO-SAM are forced to run in real-time. LIOM, on the other hand, is given infinite time to process every sensor measurement. All the methods are implemented in C++ and executed on a laptop equipped with an Intel i7-10710U CPU using the robot operating system (ROS) \cite{ROS-2009} in Ubuntu Linux. We note that only the CPU is used for computation, without parallel computing enabled. Our implementation of LIO-SAM is freely available on Github\footnote{\url{https://github.com/TixiaoShan/LIO-SAM}}. Supplementary details of the experiments performed, including complete visualizations of all tests, can be found at the link below\footnote{\url{https://youtu.be/A0H8CoORZJU}}.


\begin{figure}[ht!]
	\centering
	\subfigure[Test environment]{\includegraphics[width=.9\columnwidth]{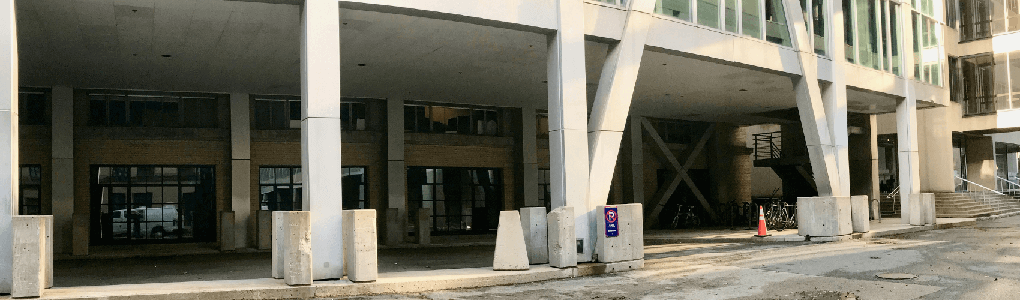}}
	\subfigure[LOAM]{\includegraphics[width=.9\columnwidth]{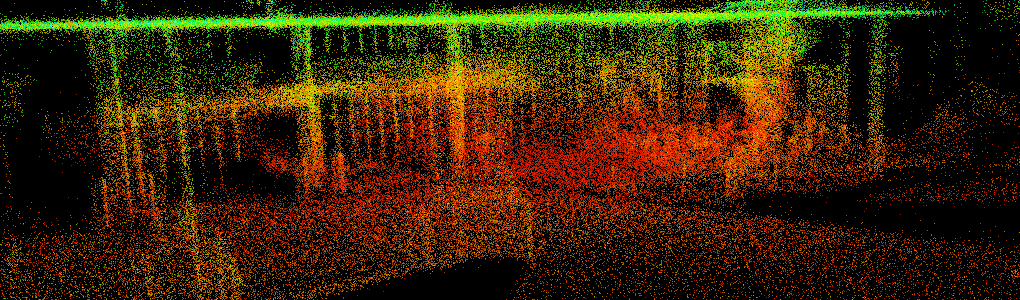}}
	\subfigure[LIO-SAM]{\includegraphics[width=.9\columnwidth]{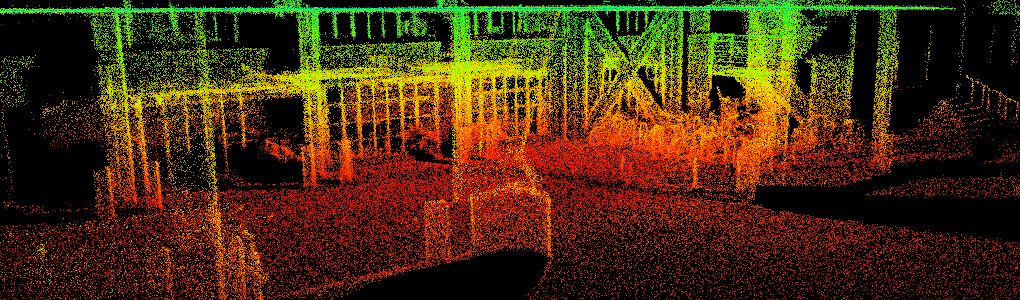}}
	\caption{Mapping results of LOAM and LIO-SAM in the \textit{Rotation} test. LIOM fails to produce meaningful results.}
	\label{fig::aggressive-rotation}
	\vspace{-3mm}
\end{figure}

\begin{figure*}[ht]
	\centering
	\subfigure[Google Earth]{\includegraphics[width=.245\textwidth]{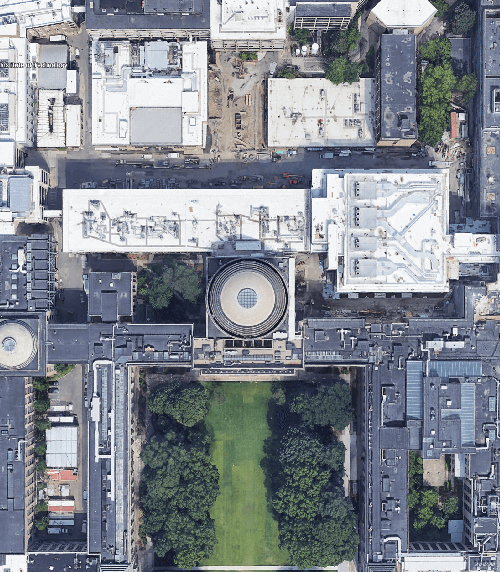}}
	\subfigure[LOAM]{\includegraphics[width=.245\textwidth]{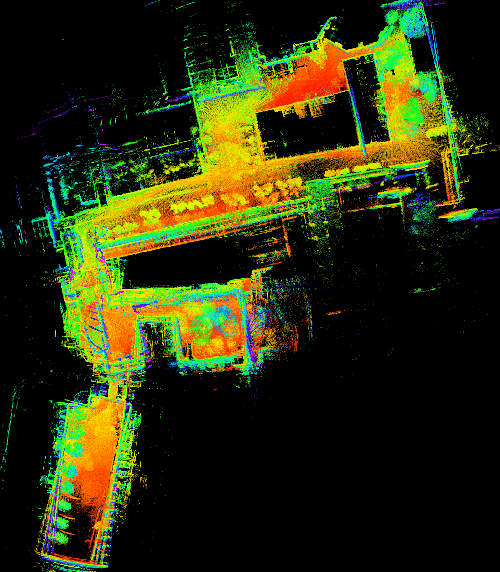}}
	\subfigure[LIOM]{\includegraphics[width=.245\textwidth]{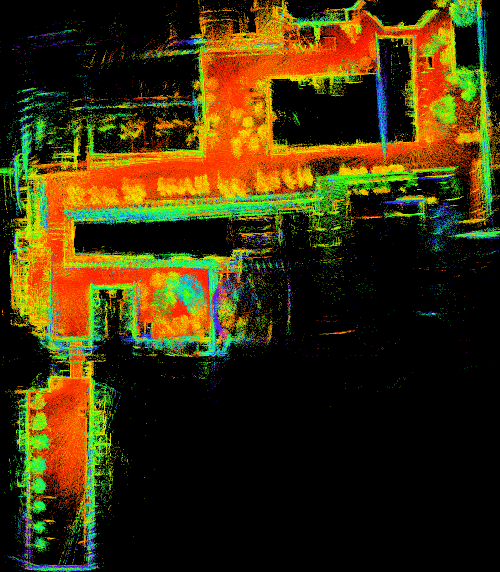}}
	\subfigure[LIO-SAM]{\includegraphics[width=.245\textwidth]{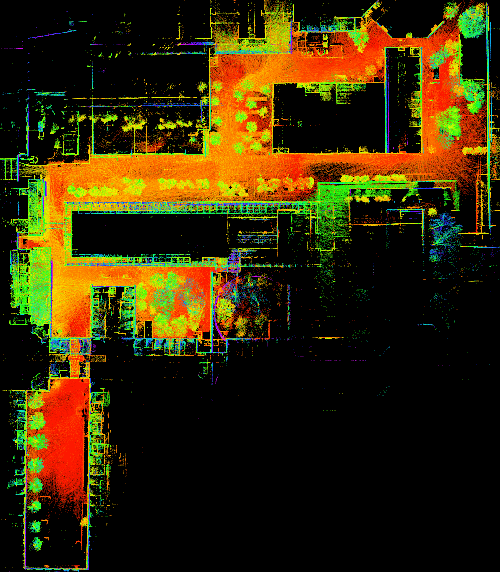}}
	\caption{Mapping results of LOAM, LIOM, and LIO-SAM using the \textit{Walking} dataset. The map of LOAM in (b) diverges multiple times when aggressive rotation is encountered. LIOM outperforms LOAM. However, its map shows numerous blurry structures due to inaccurate point cloud registration. LIO-SAM produces a map that is consistent with the Google Earth imagery, without using GPS.}
	\label{fig::aggressive-walking}
	\vspace{-5mm}
\end{figure*}

\subsection{Rotation Dataset}

In this test, we focus on evaluating the robustness of our framework when only IMU preintegration and lidar odometry factors are added to the factor graph. The \textit{Rotation} dataset is collected by a user holding the sensor suite and performing a series of aggressive rotational maneuvers while standing still. The maximum rotational speed encountered in this test is 133.7 $^{\circ}/s$. The test environment, which is populated with structures, is shown in Fig. \ref{fig::aggressive-rotation}(a). The maps obtained from LOAM and LIO-SAM are shown in Figs. \ref{fig::aggressive-rotation}(b) and (c) respectively. Because LIOM uses the same initialization pipeline from \cite{VINS-MONO}, it inherits the same initialization sensitivity of visual-inertial SLAM and is not able to initialize properly using this dataset. Due to its failure to produce meaningful results, the map of LIOM is not shown. As is shown, the map of LIO-SAM preserves more fine structural details of the environment compared with the map of LOAM. This is because LIO-SAM is able to register each lidar frame precisely in $SO(3)$, even when the robot undergoes rapid rotation. 

\subsection{Walking Dataset}

This test is designed to evaluate the performance of our method when the system undergoes aggressive translations and rotations in $SE(3)$. The maximum translational and rotational speed encountered is this dataset is 1.8 m/s and 213.9 $^{\circ}$/s respectively.
During the data gathering, the user holds the sensor suite shown in Fig. \ref{fig::experiments-equipments}(a) and walks quickly across the MIT campus (Fig. \ref{fig::aggressive-walking}(a)). In this test, the map of LOAM, shown in Fig. \ref{fig::aggressive-walking}(b), diverges at multiple locations when aggressive rotation is encountered. LIOM outperforms LOAM in this test. However, its map, shown in Fig. \ref{fig::aggressive-walking}(c), still diverges slightly in various locations and consists of numerous blurry structures. Because LIOM is designed to process all sensor measurements, it only runs at $0.56\times$ real-time while other methods are running in real-time. Finally, LIO-SAM outperforms both methods and produces a map that is consistent with the available Google Earth imagery.

\subsection{Campus Dataset}

\begin{figure}[h]
	\centering
	\subfigure[Trajectory comparison]{\includegraphics[width=.95\columnwidth]{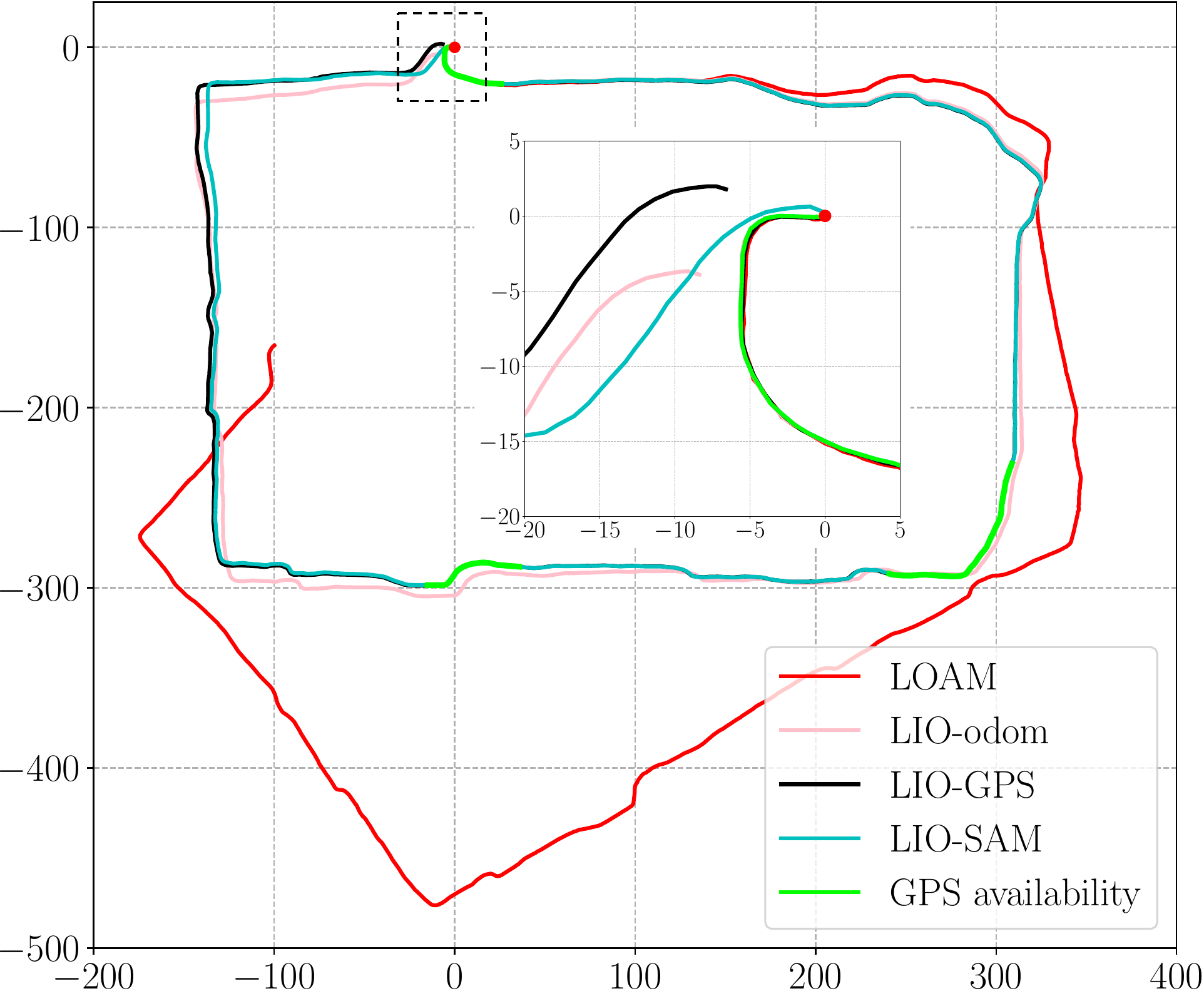}}
	\subfigure[LIO-SAM map aligned with Google Earth]{\includegraphics[width=.95\columnwidth]{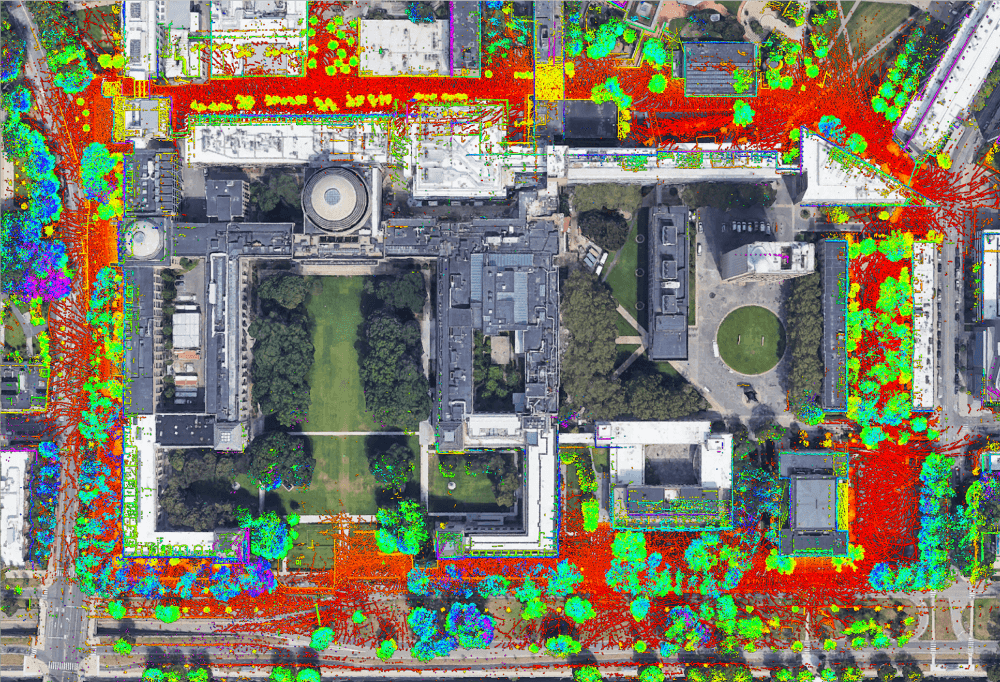}}
	\caption{Results of various methods using the \textit{Campus} dataset that is gathered on the MIT campus. The red dot indicates the start and end location. The trajectory direction is clock-wise. LIOM is not shown because it fails to produce meaningful results.}
	\label{fig::campus}
	\vspace{-1mm}
\end{figure}

\begin{table}[h]
	\centering
	\caption{End-to-end translation error (meters)}
	\label{tab::translation-error}
	\resizebox{0.95\columnwidth}{!}{
		\begin{tabular}{@{}cccccc@{}}
			\toprule
			Dataset & LOAM & LIOM & LIO-odom & LIO-GPS & LIO-SAM \\
			\midrule
			Campus & 192.43 & Fail & 9.44 & 6.87 & 0.12 \\
			Park & 121.74 & 34.60 & 36.36 & 2.93 & 0.04 \\
			Amsterdam & Fail & Fail & Fail & 1.21 & 0.17 \\
			\bottomrule
		\end{tabular}
	}
	\vspace{-3mm}
\end{table}

This test is designed to show the benefits of introducing GPS and loop closure factors. In order to do this, we purposely disable the insertion of GPS and loop closure factors into the graph. When both GPS and loop closure factors are disabled, our method is referred to as \textit{LIO-odom}, which only utilizes IMU preintegration and lidar odometry factors. When GPS factors are used, our method is referred to as \textit{LIO-GPS}, which uses IMU preintegration, lidar odometry, and GPS factors for graph construction. LIO-SAM uses all factors when they are available.

To gather this dataset, the user walks around the MIT campus using the handheld device and returns to the same position. Because of the numerous buildings and trees in the mapping area, GPS reception is rarely available and inaccurate most of the time. After filtering out the inconsistent GPS measurements, the regions where GPS is available are shown in Fig. \ref{fig::campus}(a) as green segments. These regions correspond to the few areas that are not surrounded by buildings or trees.

The estimated trajectories of LOAM, LIO-odom, LIO-GPS, and LIO-SAM are shown in Fig. \ref{fig::campus}(a). The results of LIOM are not shown due to its failure to initialize properly and produce meaningful results. As is shown, the trajectory of LOAM drifts significantly when compared with all other methods. Without the correction of GPS data, the trajectory of LIO-odom begins to visibly drift at the lower right corner of the map. With the help of GPS data, LIO-GPS can correct the drift when it is available. However, GPS data is not available in the later part of the dataset. As a result, LIO-GPS is unable to close the loop when the robot returns to the start position due to drift. On the other hand, LIO-SAM can eliminate the drift by adding loop closure factors to the graph. The map of LIO-SAM is well-aligned with Google Earth imagery and shown in Fig. \ref{fig::campus}(b). The relative translational error of all methods when the robot returns to the start is shown in Table \ref{tab::translation-error}.

\subsection{Park Dataset}

\begin{figure}[h]
	\centering
	\subfigure[Trajectory comparison]{\includegraphics[width=.95\columnwidth]{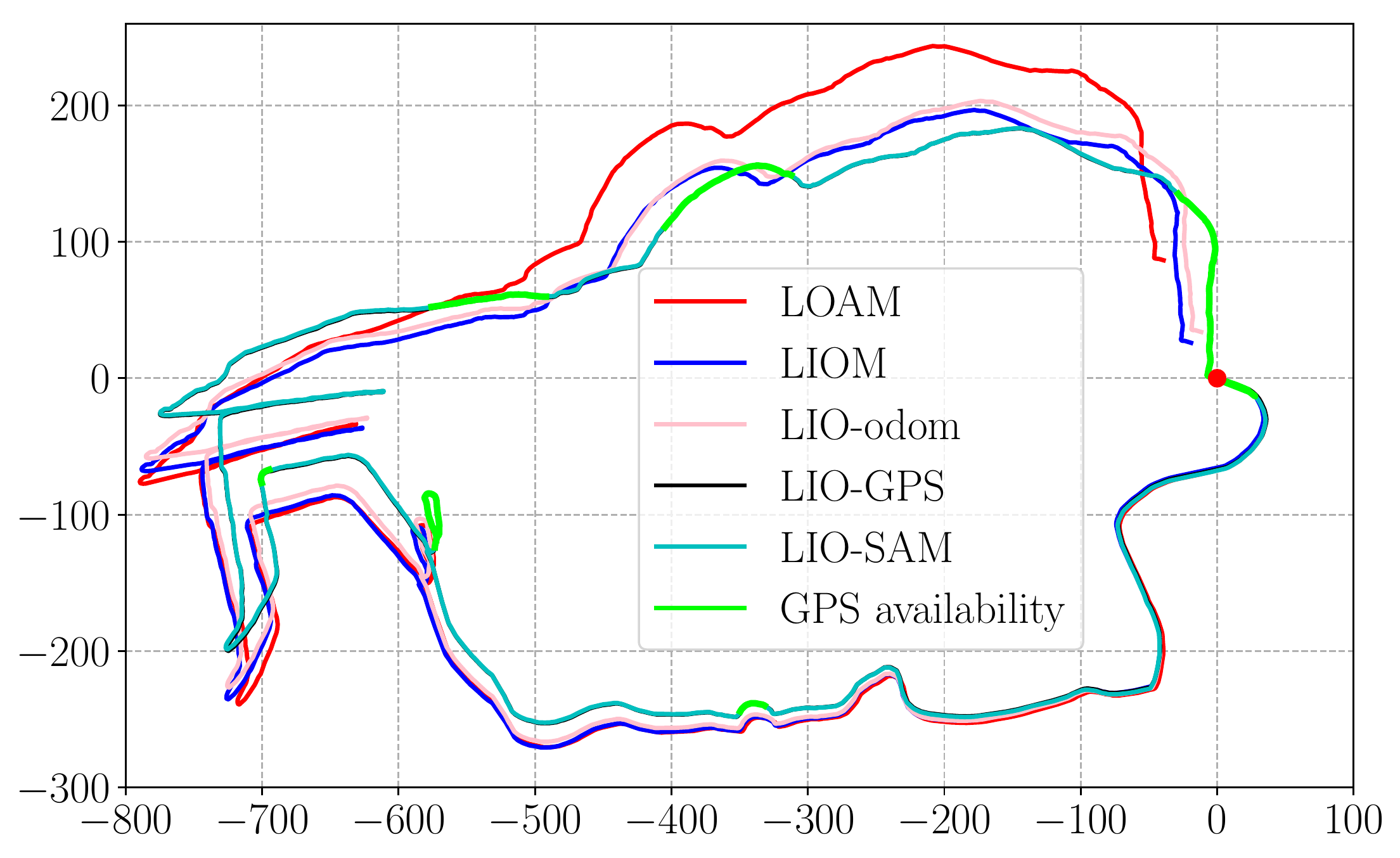}}
	\subfigure[LIO-SAM map aligned with Google Earth]{\includegraphics[width=.9\columnwidth]{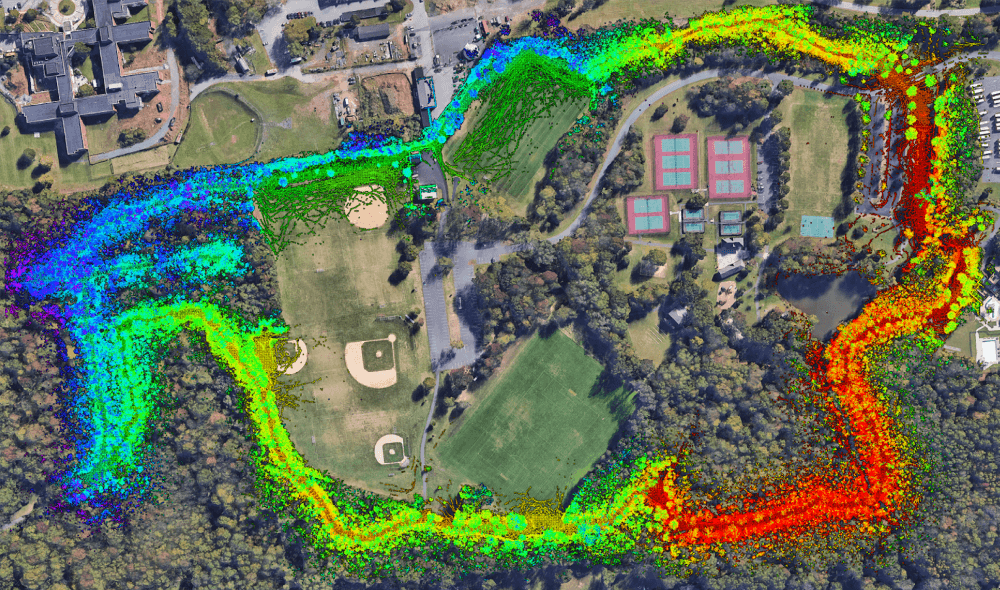}}
	\caption{Results of various methods using the \textit{Park} dataset that is gathered in Pleasant Valley Park, New Jersey. The red dot indicates the start and end location. The trajectory direction is clock-wise.  }
	\label{fig::park}
	\vspace{-3mm}
\end{figure}

In this test, we mount the sensors on a UGV and drive the vehicle along a forested hiking trail. The robot returns to its initial position after 40 minutes of driving. The UGV is driven on three road surfaces: asphalt, ground covered by grass, and dirt-covered trails. Due to its lack of suspension, the robot undergoes low amplitude but high frequency vibrations when driven on non-asphalt roads. 

To mimic a challenging mapping scenario, we only use GPS measurements when the robot is in widely open areas, which is indicated by the green segments in Fig. \ref{fig::park}(a). Such a mapping scenario is representative of a task in which a robot must map multiple GPS-denied regions and periodically returns to regions with GPS availability to correct the drift.

Similar to the results in the previous tests, LOAM, LIOM, and LIO-odom suffer from significant drift, since no absolute correction data is available. Additionally, LIOM only runs at $0.67\times$ real-time, while the other methods run in real-time. Though the trajectories of LIO-GPS and LIO-SAM coincide in the horizontal plane, their relative translational errors are different (Table \ref{tab::translation-error}). Because no reliable absolute elevation measurements are available, LIO-GPS suffers from drift in altitude and is unable to close the loop when returning to the robot's initial position. LIO-SAM has no such problem, as it utilizes loop closure factors to eliminate the drift.

\subsection{Amsterdam Dataset}

\begin{figure}[h]
	\centering
	\includegraphics[width=.9\columnwidth]{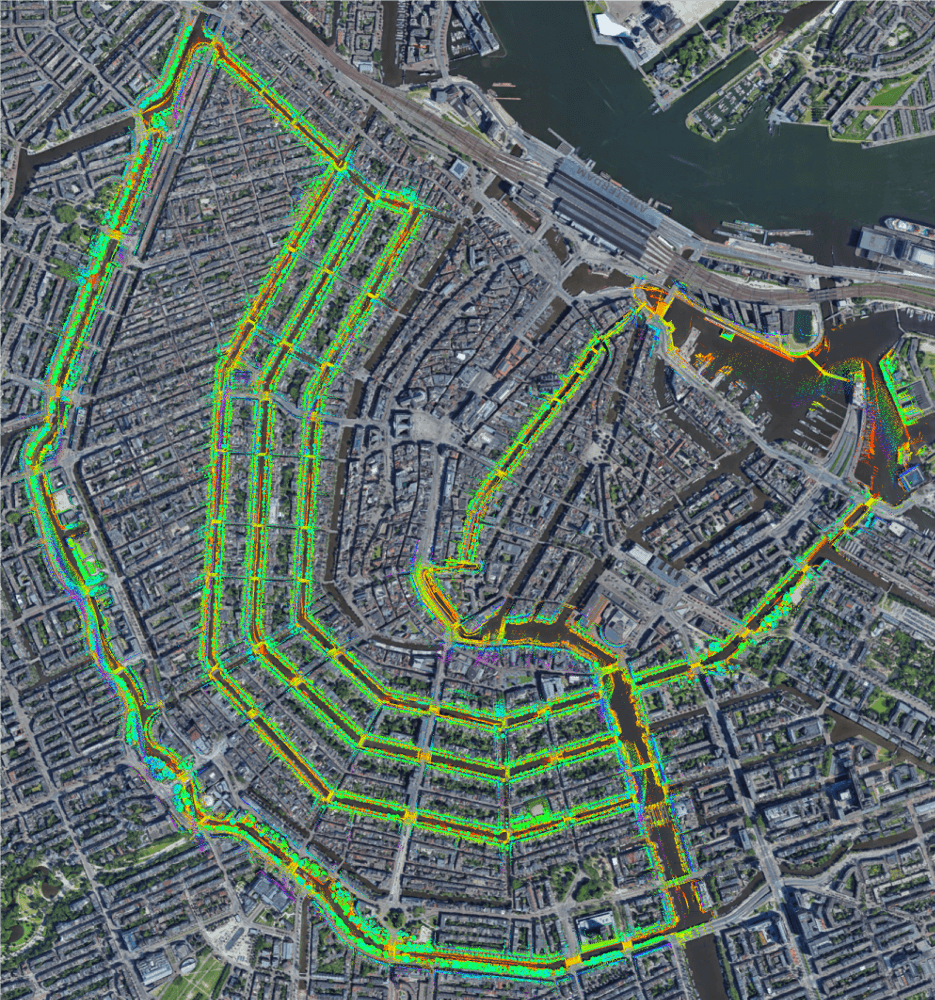}
	\caption{Map of LIO-SAM aligned with Google Earth.}
	\label{fig::amsterdam}
	\vspace{-3mm}
\end{figure}

Finally, we mounted the sensor suite on a boat and cruised along the canals of Amsterdam for 3 hours. Although the movement of the sensors is relatively smooth in this test, mapping the canals is still challenging for several reasons. Many bridges over the canals pose degenerate scenarios, as there are few useful features when the boat is under them, similar to moving through a long, featureless corridor. The number of planar features is also significantly less, as the ground is not present. We observe many false detections from the lidar when direct sunlight is in the sensor field-of-view, which occurs about 20\% of the time during data gathering. We also only obtain intermittent GPS reception due to the presence of bridges and city buildings overhead. 

Due to these challenges, LOAM, LIOM, and LIO-odom all fail to produce meaningful results in this test. Similar to the problems encountered in the \textit{Park} dataset, LIO-GPS is unable to close the loop when returning to the robot's initial position because of the drift in altitude, which further motivates our usage of loop closure factors in LIO-SAM.

\subsection{Benchmarking Results}

\begin{table}[h]
	\centering
	\caption{RMSE translation error w.r.t GPS}
	\label{tab::rmse-error}
	\resizebox{0.95\columnwidth}{!}{
		\begin{tabular}{@{}cccccc@{}}
			\toprule
			Dataset & LOAM & LIOM & LIO-odom & LIO-GPS & LIO-SAM \\
			\midrule
			Park & 47.31 & 28.96 & 23.96 & 1.09 & 0.96 \\
			\bottomrule
		\end{tabular}
	}
	\vspace{-1mm}
\end{table}

Since full GPS coverage is only available in the $Park$ dataset, we show the root mean square error (RMSE) results w.r.t to the GPS measurement history, which is treated as ground truth. This RMSE error does not take the error along the $z$ axis into account. As is shown in Table \ref{tab::rmse-error}, LIO-GPS and LIO-SAM achieve similar RMSE error with respect to the GPS ground truth. Note that we could further reduce the error of these two methods by at least an order of magnitude by giving them full access to all GPS measurements. However, full GPS access is not always available in many mapping settings. Our intention is to design a robust system that can operate in a variety of challenging environments.


\begin{table}[]
	\centering
	\caption{Runtime of mapping for processing one scan (ms)}
	\label{tab::runtime}
	\resizebox{0.9\columnwidth}{!}{
		\begin{tabular}{@{}ccccc@{}}
			\toprule
			Dataset   & LOAM  & LIOM  & LIO-SAM & Stress test \\ \midrule
			Rotation  & 83.6  & Fail  & 41.9    & 13$\times$         \\
			Walking   & 253.6 & 339.8 & 58.4    & 13$\times$         \\
			Campus    & 244.9 & Fail  & 97.8    & 10$\times$         \\
			Park      & 266.4 & 245.2 & 100.5   & 9$\times$          \\
			Amsterdam & Fail  & Fail  & 79.3    & 11$\times$         \\ \bottomrule
		\end{tabular}
	}
	\vspace{-3mm}
\end{table}

The average runtime for the three competing methods to register one lidar frame across all five datasets is shown in Table \ref{tab::runtime}. Throughout all tests, LOAM and LIO-SAM are forced to run in real-time. In other words, some lidar frames are dropped if the runtime takes more than 100$ms$ when the lidar rotation rate is 10Hz. LIOM is given infinite time to process every lidar frame. As is shown, LIO-SAM uses significantly less runtime than the other two methods, which makes it more suitable to be deployed on low-power embedded systems.

We also perform stress tests on LIO-SAM by feeding it the data faster than real-time. The maximum data playback speed is recorded and shown in the last column of Table \ref{tab::runtime} when LIO-SAM achieves similar performance without failure compared with the results when the data playback speed is 1$\times$ real-time. As is shown, LIO-SAM is able to process data faster than real-time up to 13$\times$.

We note that the runtime of LIO-SAM is more significantly influenced by the density of the feature map, and less affected by the number of nodes and factors in the factor graph. For instance, the \textit{Park} dataset is collected in a feature-rich environment where the vegetation results in a large quantity of features, whereas the \text{Amsterdam} dataset yields a sparser feature map. While the factor graph of the \text{Park} test consists of 4,573 nodes and 9,365 factors, the graph in the \text{Amsterdam} test has 23,304 nodes and 49,617 factors. Despite this, LIO-SAM uses less time in the \textit{Amsterdam} test as opposed to the runtime in the \text{Park} test.


\section{Conclusions and Discussion}

We have proposed LIO-SAM, a framework for tightly-coupled lidar inertial odometry via smoothing and mapping, for performing real-time state estimation and mapping in complex environments. By formulating lidar-inertial odometry atop a factor graph, LIO-SAM is especially suitable for multi-sensor fusion. Additional sensor measurements can easily be incorporated into the framework as new factors. Sensors that provide absolute measurements, such as a GPS, compass, or altimeter, can be used to eliminate the drift of lidar inertial odometry that accumulates over long durations, or in feature-poor environments. Place recognition can also be easily incorporated into the system. To improve the real-time performance of the system, we propose a sliding window approach that marginalizes old lidar frames for scan-matching. Keyframes are selectively added to the factor graph, and new keyframes are registered only to a fixed-size set of sub-keyframes when both lidar odometry and loop closure factors are generated. This scan-matching at a local scale rather than a global scale facilitates the real-time performance of the LIO-SAM framework. The proposed method is thoroughly evaluated on datasets gathered on three platforms across a variety of environments. The results show that LIO-SAM can achieve similar or better accuracy when compared with LOAM and LIOM. Future work involves testing the proposed system on unmanned aerial vehicles.


%


\vspace{-1mm}

\begin{thebibliography}{99}


\bibitem{LOAM-2017}
J. Zhang and S. Singh, ``Low-drift and Real-time Lidar Odometry and Mapping," \textit{Autonomous Robots}, vol. 41(2): 401-416, 2017.

\bibitem{KITTI-2012}
A. Geiger, P. Lenz, and R. Urtasun, ``Are We Ready for Autonomous Driving? The KITTI Vision Benchmark Suite", \textit{IEEE International Conference on Computer Vision and Pattern Recognition}, pp. 3354-3361, 2012.

\bibitem{ICP-1992}
P.J. Besl and N.D. McKay, ``A Method for Registration of 3D Shapes," \textit{IEEE Transactions on Pattern Analysis and Machine Intelligence}, vol. 14(2): 239-256, 1992.

\bibitem{GICP-2009}
A. Segal, D. Haehnel, and S. Thrun, ``Generalized-ICP," \textit{Proceedings of Robotics: Science and Systems}, 2009.

\bibitem{Finding-Plane-2013}
W.S. Grant, R.C. Voorhies, and L. Itti, ``Finding Planes in LiDAR Point Clouds for Real-time Registration," \textit{IEEE/RSJ International Conference on Intelligent Robots and Systems}, pp. 4347-4354, 2013.

\bibitem{Collar-Line-2016}
M. Velas, M. Spanel, and A. Herout, ``Collar Line Segments for Fast Odometry Estimation from Velodyne Point Clouds," \textit{IEEE International Conference on Robotics and Automation}, pp. 4486-4495, 2016.

\bibitem{LEGOLOAM-2018}
T. Shan and B. Englot, ``LeGO-LOAM: Lightweight and Ground-optimized Lidar Odometry and Mapping on Variable Terrain," \textit{IEEE/RSJ International Conference on Intelligent Robots and Systems}, pp. 4758-4765, 2018.

\bibitem{Traversability-2018}
T. Shan, J. Wang, K. Doherty, and B. Englot, ``Bayesian Generalized Kernel Inference for Terrain Traversability Mapping," \textit{In Conference on Robot Learning}, pp. 829-838, 2018.

\bibitem{UAV-IROS-2013}
S. Lynen, M.W. Achtelik, S. Weiss, M. Chli, and R. Siegwart, ``A Robust and Modular Multi-sensor Fusion Approach Applied to MAV Navigation," \textit{IEEE/RSJ International Conference on Intelligent Robots and Systems}, pp. 3923-3929, 2013.

\bibitem{DiDi-Mapping-2018}
S. Yang, X. Zhu, X. Nian, L. Feng, X. Qu, and T. Mal, ``A Robust Pose Graph Approach for City Scale LiDAR Mapping," \textit{IEEE/RSJ International Conference on Intelligent Robots and Systems}, pp. 1175-1182, 2018.

\bibitem{Honda-2019}
M. Demir and K. Fujimura, ``Robust Localization with Low-Mounted Multiple LiDARs in Urban Environments," \textit{IEEE Intelligent Transportation Systems Conference}, pp. 3288-3293, 2019.

\bibitem{Gao-Sensors-2015}
Y. Gao, S. Liu, M. Atia, and A. Noureldin, ``INS/GPS/LiDAR Integrated Navigation System for Urban and Indoor Environments using Hybrid Scan Matching Algorithm," \textit{Sensors}, vol. 15(9): 23286-23302, 2015.

\bibitem{Hening-AIAA-2017}
S. Hening, C.A. Ippolito, K.S. Krishnakumar, V. Stepanyan, and M. Teodorescu, ``3D LiDAR SLAM integration with GPS/INS for UAVs in urban GPS-degraded environments," \textit{AIAA Infotech@Aerospace Conference}, pp. 448-457, 2017.

\bibitem{SLAM-Review-2018}
C. Chen, H. Zhu, M. Li, and S. You, ``A Review of Visual-Inertial Simultaneous Localization and Mapping from Filtering-Based and Optimization-Based Perspectives," \textit{Robotics}, vol. 7(3):45, 2018.


\bibitem{IN2LAMA-ICRA-2019}
C. Le Gentil,, T. Vidal-Calleja, and S. Huang, ``IN2LAMA: Inertial Lidar Localisation and Mapping," \textit{IEEE International Conference on Robotics and Automation}, pp. 6388-6394, 2019.

\bibitem{Robocentric-2019}
C. Qin, H. Ye, C.E. Pranata, J. Han, S. Zhang, and Ming Liu, ``R-LINS: A Robocentric Lidar-Inertial State Estimator for Robust and Efficient Navigation," \textit{arXiv:1907.02233}, 2019.

\bibitem{LIO-MAPPING-2019}
H. Ye, Y. Chen, and M. Liu, ``Tightly Coupled 3D Lidar Inertial Odometry and Mapping," \textit{IEEE International Conference on Robotics and Automation}, pp. 3144-3150, 2019.

\bibitem{Dellaert-Factor-Graph-2017}
F. Dellaert and M. Kaess, ``Factor Graphs for Robot Perception," \textit{Foundations and Trends in Robotics}, vol. 6(1-2): 1-139, 2017.

\bibitem{iSAM2-2012}
M. Kaess, H. Johannsson, R. Roberts, V. Ila, J.J. Leonard, and F. Dellaert, ``iSAM2: Incremental Smoothing and Mapping Using the Bayes Tree," \textit{The International Journal of Robotics Research}, vol. 31(2): 216-235, 2012.

\bibitem{IMUPre-2016}
C. Forster, L. Carlone, F. Dellaert, and D. Scaramuzza, ``On-Manifold Preintegration for Real-Time Visual-Inertial Odometry," \textit{IEEE Transactions on Robotics}, vol. 33(1): 1-21, 2016.

\bibitem{ROBOT-LOCALIZATION}
T. Moore and D. Stouch, ``A Generalized Extended Kalman Filter Implementation for The Robot Operating System," \textit{Intelligent Autonomous Systems}, vol. 13: 335-348, 2016.


\bibitem{LOOP-CLOSURE-2018}
G. Kim and A. Kim, ``Scan Context: Egocentric Spatial Descriptor for Place Recognition within 3D Point Cloud Map," \textit{IEEE/RSJ International Conference on Intelligent Robots and Systems}, pp. 4802-4809, 2018.

\bibitem{LOOP-CLOSURE-2019}
J. Guo, P. VK Borges, C. Park, and A. Gawel, ``Local Descriptor for Robust Place Recognition using Lidar Intensity," \textit{IEEE Robotics and Automation Letters}, vol. 4(2): 1470-1477, 2019.

\bibitem{ROS-2009}
M. Quigley, K. Conley, B. Gerkey, J. Faust, T. Foote, J. Leibs, R. Wheeler, and A.Y. Ng, ``ROS: An Open-source Robot Operating System," \textit{IEEE ICRA Workshop on Open Source Software}, 2009.

\bibitem{VINS-MONO}
T. Qin, P. Li, and S. Shen, ``Vins-mono: A Robust and Versatile Monocular Visual-Inertial State Estimator," \textit{IEEE Transactions on Robotics}, vol. 34(4): 1004-1020, 2018.

\end{thebibliography}
\end{document}